# Evolution of a Subsumption Architecture Neurocontroller

Julian Togelius

Mäster Nilsgatan 2
211 26 Malmö
Sweden
+46-705-192088
julian@togelius.com

**Abstract.** An approach to robotics called layered evolution and merging features from the subsumption architecture into evolutionary robotics is presented, and its advantages are discussed. This approach is used to construct a layered controller for a simulated robot that learns which light source to approach in an environment with obstacles. The evolvability and performance of layered evolution on this task is compared to (standard) monolithic evolution, incremental and modularised evolution. To corroborate the hypothesis that a layered controller performs at least as well as an integrated one, the evolved layers are merged back into a single network. On the grounds of the test results, it is argued that layered evolution provides a superior approach for many tasks, and it is suggested that this approach may be the key to scaling up evolutionary robotics.

## 1. Introduction: Layered Evolution

One of the chief problems of evolutionary robotics is scaling up; that is, evolving robot controllers that solve complicated problems and tackle complicated, variable environments. Many strategies to achieve scaling up have been devised and tried, of which two of the most prominent are incremental evolution and modularised evolution.

Incremental evolution refers to the practice of changing the fitness function during the evolution of the controller, e.g. a coke can-collecting robot might be evolved with its fitness initially dependent on how close on average it gets to red objects, and when the mean population fitness is high enough the fitness function is changed so that fitness is made dependent on how many cans are actually collected [10] [17]. Evolvability is thus increased by smoothing the fitness landscape of the evolutionary algorithm; if the latter criterion (number of cans picked up) was used from the beginning, the fitness landscape would probably be too steep to allow any progress, as very few if any in a random population of neurocontrollers would pick up any cans at all.

Modularised evolution is my term for work performed by a number of different authors, for example those in Stefano Nolfi's research team, who evolve robot controllers consisting of more than one neural network [7] [12]. In some experiments several networks are present from the start, and in others one network is evolved and then duplicated, but in none of these experiments functional differentiation between the networks is decided by humans. (Though this has been explored in other contexts and using backpropagation rather than evolution [18].)

I believe these two strategies can be combined, and that in doing so there is much to learn from looking at neighbouring research fields. Behaviour-based robotics is an active research field since the mid-eighties, when Rodney Brooks invented the subsumption architecture, upon variations of which the majority of work in the field is done [4] [5]. The relative complexity of tasks solvable by robots in this tradition merits a closer look at key features of the paradigm. A subsumption architecture robot controller is divided into layers, where each layer is a human-designed piece of hardware or software responsible for a particular behaviour. Lower layers are responsible for the simpler and more vital behaviours, and higher layers for more complicated behaviours. The chain of command is strictly unidirectional: higher layers are allowed to influence and ovveride (subsume) the outputs of lower layers, and are allowed to dependent on lower layers for their functioning; lower layers are never allowed to depend on higher layers being there in order for them to function properly. In a typical robot, most layers receive inputs from sensors, while only the lower layers directly influence actuators.

The approach to evolutionary robotics I am advocating and pursuing in this paper, Layered evolution, is combining incremental and modularised evolution with elements from the subsumption architecture. Basically, a robot in layered evolution is controlled by a subsumption-style controller where each layer is an evolutionary neural network of one of several possible types. (Evolution of subsumption controllers was proposed by Brooks in 1991, but his idea was to evolve program code rather than neural networks; as far as I know, no experiments in evolving neural layers have been published. [6]) The neural network-layers are evolved in sequence, so that the lowest layer is evolved first, and only after desired fitness for that layer is reached, its development is stopped, another layer is added on top of it, the fitness function is changed, and evolution of the new layer commences. In the evolution of any one layer, all lower layers are present and frozen (kept fixed), and their configuration is identical in all individuals in the population undergoing evolution; only the configuration of the highest layer differs between individuals.

The relation between vintage evolutionary robotics (subsequently referred to as monolithic evolution, as it only uses one network in the controller), incremental and modularised evolution is illustrated in [figure 1].

|  | One layer | Many layers |
|---|---|---|
| One fitness function | Monolithic evolution | Modularised evolution |
| Many fitness functions | Incremental evolution | Layered evolution |

**Fig. 1.** Four different approaches to evolutionary robotics.

## 2. The advantages of layered evolution

These are the reasons I hypothesize layered evolution faster and more reliably produces better solutions to a given problem than the standard approach does:

**2.1. Beyond incremental evolution**
To begin with, advantages of incremental evolution carry over into layered evolution. As the separate layers perform different parts of the task, they must be evolved with different fitness criteria, as is the case with incremental evolution, and this smoothens the fitness landscape.

**2.2. Network size: updating speed and search dimensionality**
As the complexity of an operation should be measured in the number of times its most frequent operation is performed, when it comes to updating neural networks the number of synapses is what counts [2]. Also, in direct encodings, each synapse occupies one position on the genome string and thus contributes one dimension to the evolutionary search space, and reducing dimensionality is essential in many classic learning algorithms [1]. Number of synapses is $O(n^2)$ (where n is the number of neurons) in most interesting network topologies. Separating the networks should theoretically be able to do wonders with scalability, as 10 networks with 10 neurons each has ten times fewer synapses than one network with 100 neurons, and this difference only gets greater as one moves towards the complexity of real nervous systems.

**2.3. Forcing functional separation helps**
By not evolving a completed mechanism anymore when starting to evolve a new one, evolution benefits in several ways. Not only are computations not wasted on trying new solutions for something that's already finished, the risk of destroying the completed mechanism, in case the search runs into a flat area where most behavioural differences do not affect fitness, is also eliminated. (When training modular networks with backpropagation, this is termed "unlearning" [18].) Apart from these more or less obvious advantages, one may consider Nolfi's recent observation that he could actually improve evolvability of a fully connected network with a strategic lesion [13]. More generally, the evolution of new mechanisms benefits by hindering old mechanisms from obstructing them, but in a fully connected network every neuron affects every other and any change *ceteris paribus* lowers fitness (neutrality is low).

### 2.4. Several network types in one controller

As noted above, several different types of neural networks are used in evolutionary robotics research. Although some researchers use networks incorporating features from several network types, adding features to a network adds to both network updating time and dimensionality of the search space [8]. A layered architecture, on the other hand, makes it easy to select the right type of network for each layer – perceptron-like for reactive behaviour, plastic for learning behaviour etcetera.

### 2.5. Good design principles and reusability

Finally, what I after all consider the main engineering argument for layered evolution is that it introduces some good engineering principles into layered evolution. Though evolution is very good at coming up with creative and indeed surprising solutions to many problems, these solutions often look like a mess to a schooled engineer [11]. Which might be why they do not scale up; only simplistic behaviours are ever evolved. Layered evolution introduces modularity into evolutionary robotics, and with modularity comes reusability. Some time in the modular future there might even be a universal repository of ready-evolved layers, free for anyone to include in his or her controller architecture and build upon.

### 2.6. Interpretations and objections

As for the scientific side of evolutionary robotics, the subsumption architecture has been interpreted in terms of neurophysiologic and behavioural layers, and inspired hypotheses in those fields. [14]. I discuss possible uses of layered evolution in scientific modelling in my MSc Dissertation, in which I also respond to some possible criticisms of the approach.

### 3. First experiment: conditional phototaxis with obstacles and changing goals

To test my main hypothesis that layered evolution faster and more reliably produces desired behaviours than monolithic and incremental evolution, I compared these methodologies when applied to the problem of designing a neural controller for a simulated robot that would perform goal-seeking behaviour and lifetime learning in a cluttered environment.

This task is interesting because it requires some quite different behaviours, from the simple reflex-like obstacle avoidance, via the less predictable but still possibly reactive conditional phototaxis to the non-reactive learning behaviour. It is also interesting because learning behaviours are usually quite difficult to evolve, and as far as I know it has not been done in an environment with obstacles [3] [16].

### 3.1. Methods

The robot simulation is not based on any existing robot, though loosely inspired by the Khepera. The robot is circular and moves about (by turning its two independently moving wheels) in a rectangular world with fixed dimensions. This world includes two light sources, and possibly also ten rectangular obstacles, all randomly placed. Two light sensor mounted in the robot's movement direction correspond to one light source each, and signal with an intensity dependent on the angle and distance to its respective light source. Three obstacle sensors (deviation –1, 0, and 1 radian from movement direction) signal when close to obstacles; running into an obstacle stops the robot's motion.

The robot is controlled by an architecture incorporating one, two or three feedforward neural networks. In the case of a single network, that network's outputs drive the wheels directly; in the other cases a subsumption mechanism connects the outputs of layer one or two to the wheels thus: if the third output of layer 2 is higher than 0,5 the first two outputs of that layer is connected to the wheels, otherwise the outputs from layer 1 is used.

All neurons use the tanh squashing function. Synapses can be either fixed, in which case a number [-2..2] per synapse is encoded directly on the genome, or plastic (activity-dependent). Updating rules and encoding for plastic synapses are taken directly from the work of Floreano and Mondada [9]. Neural networks can consist of all fixed, all plastic, or a combination of both synapse types. All networks have an extra input set to the constant value one.

For evolving the networks, an algorithm where the least fit half of the 30 genome population was replaced with clones of the other half and all genomes except the five best were mutated every generation, was used. Each genome's fitness was calculated as the lowest fitness of five trials. This was done for 100 generations, 10 times for each task.

The robot's ultimate task can be divided into three subtasks: conditional phototaxis, which means moving and staying close to the correct light source as specified by an external input, obstacle avoidance, and learning. Fitness for the first two subtasks is measured as mean distance to the correct light source over 200 time steps, with or without obstacles. In the learning subtask, the robot is not told which light source is the right one (varies randomly between trials), but only gets a feedback on which light source it has touched and whether that was correct upon touching it, at which time both sources are moved to random locations. Fitness is measured as number of times the right light source was touched minus number of times the wrong one was touched when the robot moved for 400 time steps.

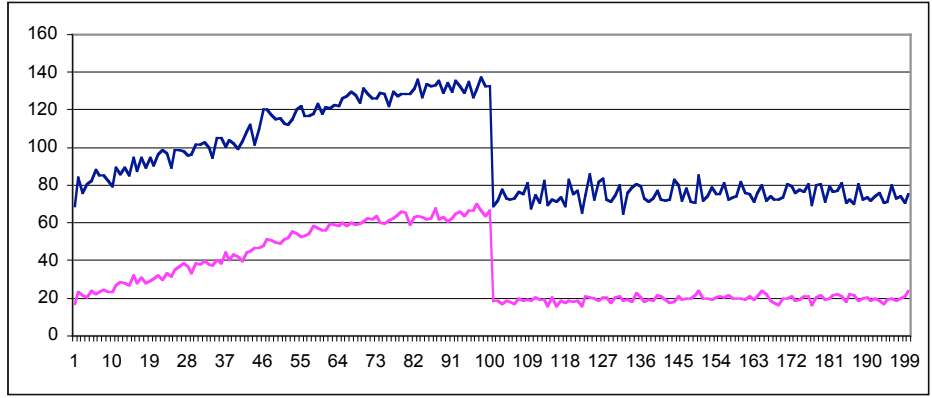

**Fig. 2.** Incremental evolution of conditional phototaxis and obstacle avoidance. Fitness (in mean distance from target light source) runs along the vertical axis and evolutionary time (in number of generations) along the horizontal axis. The upper line shows the fitness of the best genome in the population in every generation, and the lower line shows the average fitness of the population. Fitness scores are averaged over ten runs of the sub-experiment. These conventions apply to all subsequent graphs in the paper.

**3.2. Results**

*3.2.1. Monolithic evolution* One network with nine inputs (two light sensors, three obstacle sensors, two light source contact sensors and one feedback, as described above), six interneurons, two outputs and all hybrid (either fixed or plastic) synapses was evolved on the full task. Mean population fitness never differed significantly from zero. To help monolithic evolution on it's a way a bit, the distribution of target light sources was changed so that in two of three cases the target was the red light source [16]. Under this condition, most evolved robots displayed rudimentary phototaxis towards the red light, but neither obstacle avoidance nor learning. Fitness stayed low.

Removing the learning element, the same network was then evolved to do just conditional phototaxis; with obstacles present, fitness stayed at chance level, but without obstacles, satisfactory phototaxis did evolve.

*3.2.2. Incremental evolution* The network mentioned above was evolved for 100 generations without obstacles, after which obstacles were added and evolution continued for 100 generations more. Conditional phototaxis evolved as quickly as before, but when obstacles were added fitness fell drastically and remained low (see [figure 2]). Remarkably, the controllers from generation 199 performed much worse when tested in an environment without obstacles than did those in generation 99; not only did obstacle avoidance not evolve, conditional phototaxis was also lost, probably due to genetic drift in the absence of selection pressure.

*3.2.3. Modularised evolution* A controller with two networks was evolved to do conditional phototaxis and obstacle avoidance. The first network had four inputs (light sources, conditional and stable), three interneurons and two outputs; the second network had three inputs (obstacle sensors and stable) and three outputs (motor outputs and subsumption). Both networks had fixed synapses were evolved simultaneously. Most controllers did good phototaxis and reasonable obstacle avoidance.

To see if this approach would work for learning as well, a third layer was added with four inputs (feedback, light source contact sensors and stable), two interneurons and one output connected to the conditional output of layer one, and all networks were evolved from scratch simultaneously. No systematic fitness increase was seen, nor was any sensible behaviour.

*3.2.4. Layered evolution* Layers were evolved sequentially, building upon each other, and the ready-evolved layers were frozen. The same networks as in 3.2.3 were used.

First, conditional phototaxis was evolved. Rather quickly, good performance was reached (see [figure 3]).

Secondly, an obstacle avoidance network was added, and evolution continued with the first layer frozen. Decent obstacle avoidance evolved in just a few generations; the only systematic mean fitness gain in [figure 4] is within the first five generations. Controllers with an evolved obstacle avoidance layer did much better in than controllers with just one layer in environments with obstacles.

Thirdly, a learning network was added. After just five generations, controllers with good learning capability were present, and after thirty generations no substantial fitness increases were seen (see [figure 5]). This suggests that learning taken for itself is not a very hard behaviour to evolve. An analysis of an evolved network confirms this suggestion. In one controller the following was the case: in the default mode, there was a covariance negative connection between the reward input and one of the interneurons, another between that interneuron and the output neuron, and a plain Hebbian connection between the stable input and the same interneuron. So the top layer always outputted zero, but if the robot ever touched the wrong light, it switched to always outputting one. This is very different and and a far bit simpler than it would have been had I designed the mechanism.

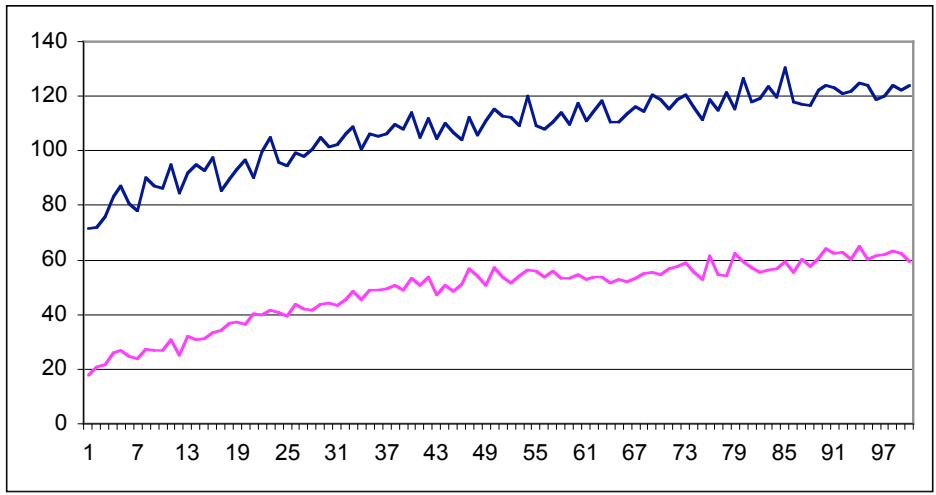

**Fig. 3.** Evolution of conditional phototaxis

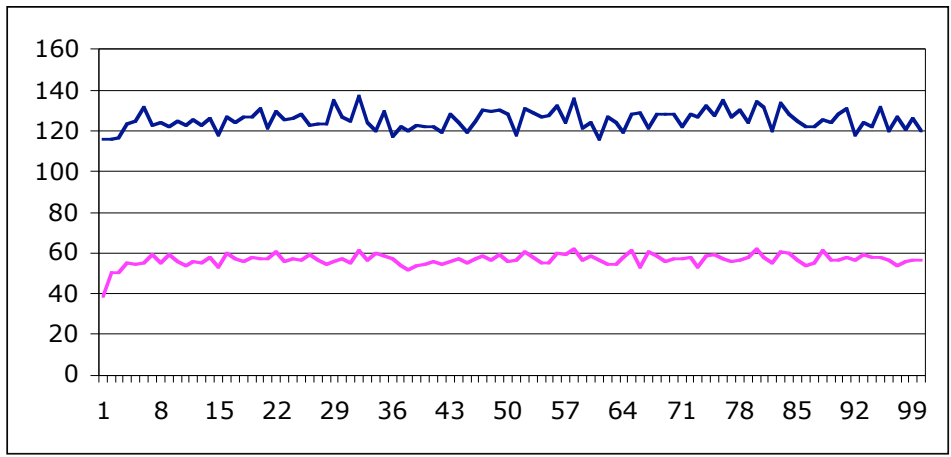

**Fig. 4.** Evolution of obstacle avoidance

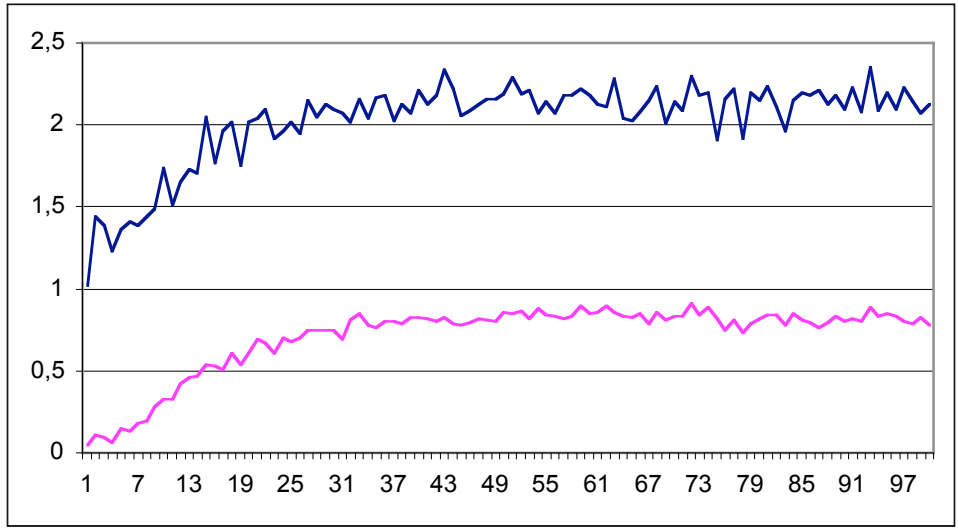

**Fig. 5.** Evolution of learning

## 4. Second experiment: merging layers

While it certainly seems that layered evolution quickly and reliably evolves desired behaviours, it cannot be excluded that there are any advantages to integrating these behaviours into one network. As a first investigation into this topic, I decided to merge together the layers of a ready-evolved layered architecture, and see if this would increase fitness beyond that of the pure layered structure.

### 4.1. Methods
To ready-evolved controllers, neural connection layers were added capable of propagating activations like fixed synapses, but from higher to lower layers. These layers had zero connections at the outset of an evolutionary run, but every generation that a particular genome was mutated, every connection in its connection layers had probability = mutation rate of being deleted; also, every layer had probability = 5 * mutation rate of adding a new connection from a random neuron in the layer above to a random neuron below, strength [-2..2].
   In the absence of selection pressure, such a layer would thus contain on average five neurons.

### 4.2. Results
The best genomes from the results of each of the 10 evolutionary runs of sub-experiment 3.2.4 were used as seed for ten new runs. First, the controllers were evolved for 100 generations more, with all networks mutable. A small fall in mean population fitness was observed within the first few generations, probably due to loss of homogeneity in the population, but after that mean fitness stood still.
   Repeating this sub-experiment with added connection layers showed no discernible difference from the first condition; mean fitness never exceeded that of the first generation. The mean size of the connection layers was three connections, which indicates some selection pressure against interlayer connections other than the hard-coded ones.
   These results may be taken as a weak indication that there are no fitness benefits to be reaped from combining these behaviours into one neural network.

## 5. Conclusion

The results presented in this paper show that for the particular robot task of doing conditional phototaxis with changing goals in an environment with obstacles, evolvability was drastically improved both by organizing the controller according to the subsumption architecture, and by evolving the layers one at a time. In fact, no other evolutionary robotics approach seems to be able to evolve a solution to the complete task, and any advantages from combining the behaviours into one network has yet to be found.

Whether the enhanced evolvability of layered evolution is truly generalizable and represents something of a golden road to scaling up for evolutionary robotics, is certainly an interesting question for the future. It is possible that many interesting problems are not amenable to behavioural deconstruction in this fashion (but finding out which ones are and which ones aren't could well tell us something interesting), and it is possible that some behavioural layers are practically impossible to evolve in themselves. This needs to be investigated.

## 6. Acknowledgements

This article is based on my MSc dissertation, which was done at the University of Sussex under supervision of Ezequiel Di Paolo (and which contains a much more detailed discussion of both theory, implementation and results of the project). Thanks to Dyre Bjerknes, Tom Ziemke and anonymous reviewers for valuable comments.